\documentclass[10pt,journal,compsoc]{IEEEtran}

\ifCLASSOPTIONcompsoc
  \usepackage[nocompress]{cite}
\else
  \usepackage{cite}
\fi

\usepackage{amssymb}
\usepackage{amsmath}
\usepackage{amsthm}
\usepackage{graphicx}
\usepackage{times}
\usepackage{epsfig}
\usepackage{subcaption}
\usepackage{booktabs}
\usepackage{hyperref}
\hypersetup{pagebackref=true,breaklinks=true,letterpaper=true,colorlinks,bookmarks=false}

\newcommand{\minisection}[1]{\vspace{2mm}\noindent{\textbf{#1}.}}

\begin{document}
\title{DeepFake Detection Based on Discrepancies Between Faces and their Context}
\author{Yuval~Nirkin, Lior Wolf, Yosi~Keller, and~Tal~Hassner~%
\IEEEcompsocitemizethanks{\IEEEcompsocthanksitem Y. Nirkin and Y. Keller are with the Faculty of Engineering. Bar Ilan University. \protect \and
E-mail: yuval.nirkin@gmail.com
\IEEEcompsocthanksitem Lior Wolf is with Facebook and Tel Aviv University.
\IEEEcompsocthanksitem Tal Hassner is with Facebook AI.
\IEEEcompsocthanksitem Work done in academia and outside FB.}
}

\IEEEtitleabstractindextext{%
\begin{abstract}
We propose a method for detecting face swapping and other identity manipulations in single images. Face swapping methods, such as DeepFake, manipulate the face region, aiming to adjust the face to the appearance of its context, while leaving the context unchanged. We show that this modus operandi produces discrepancies between the two regions (e.g., Fig.~\ref{fig:teaser}). These discrepancies offer exploitable telltale signs of manipulation. Our approach involves two networks: (i) a face identification network that considers the face region bounded by a tight semantic segmentation, and (ii) a context recognition network that considers the face context (e.g., hair, ears, neck). We describe a method which uses the recognition signals from our two networks to detect such discrepancies, providing a complementary detection signal that improves conventional real vs. fake classifiers commonly used for detecting fake images. Our method achieves state of the art results on the FaceForensics++, Celeb-DF-v2, and DFDC benchmarks for face manipulation detection, and even generalizes to detect fakes produced by unseen methods.
\end{abstract}

}

\maketitle

\IEEEdisplaynontitleabstractindextext
\IEEEpeerreviewmaketitle

\IEEEraisesectionheading{\section{Introduction}\label{sec:introduction}}

\begin{figure*}[!htb]
\centering
\includegraphics[width=1.0\textwidth]{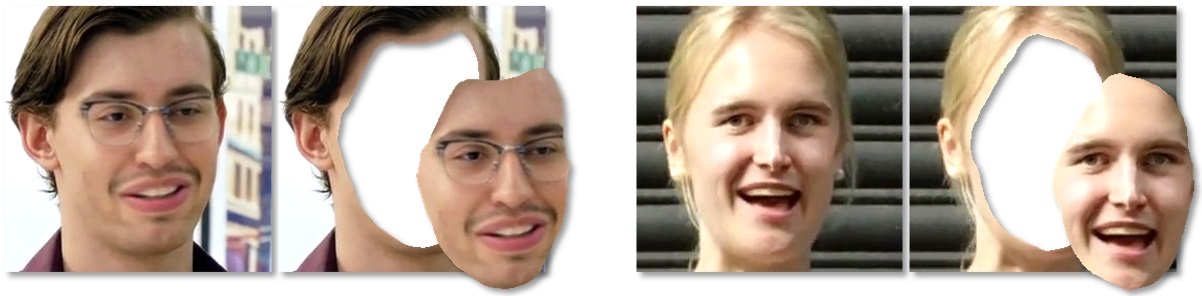}
\caption{{\bf Detecting swapped faces by comparing faces and their context.} Two example fake (swapped) faces from DFD~\cite{dfd}. Left: The arm of the eyeglasses does not extend from face to context. Right: An apparent identity mismatch between face and context. We show how these and similar discrepancies can be used as powerful signals for automatic detection of swapped faces.}\vspace{2mm}
\label{fig:teaser}
\end{figure*}

\IEEEPARstart{P}{hotography} is widely perceived as offering authentic evidence of actual
events, including, in particular, the presence and actions of human subjects
in images and videos. Although this perception is slowly shifting,
contemporary technology allows far easier and more accessible manipulation
of images than many realize. This gap represents a societal threat
whenever manipulated media is released over social networks and consumed by
a public that is ill-equipped to question its authenticity.

For instance, existing technology makes it easier for an actor to speak a
given text, and then change her facial appearance and voice to imitate those
of someone else. Alternatively, the face of a person captured in a
crime-scene can be manipulated and replaced by another. Both of these examples are
referred to as \emph{face swapping}. A third scenario involves the 
reenactment of a person's face to change expression or lip motion (aka \emph{%
face reenactment}).

\begin{figure*}[!htb]
\centering
\includegraphics[width=1.0\linewidth]{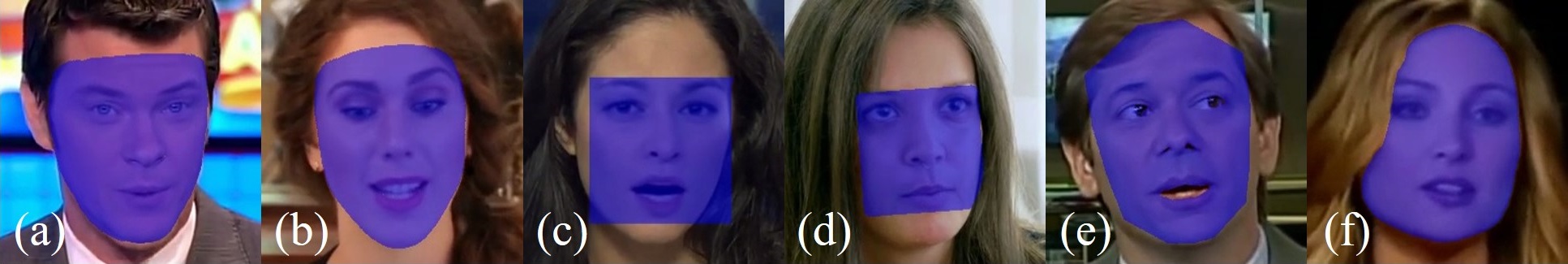}
\caption{\textbf{Affected regions of different manipulation methods}. (a) + (b) Face2Face~\cite{thies2016face2face} and NeuralTextures~\cite{thies2019deferred}; (c) + (d) Deepfake~\cite{DeepFakes} variants of FaceForensics++~\cite{roessler2019faceforensics++} and DFD~\cite{dfd}; (e) FaceSwap~\cite{FaceSwap}; (f) FSGAN~\cite{nirkin2019fsgan}. In all cases, faces are manipulated but their context is left unchanged.\vspace{-3mm}}
\label{fig:manipulated_regions}
\end{figure*}

Contemporary approaches for detecting such manipulations relate to these
three scenarios similarly: by training a classifier to distinguish between
real and fake images or videos~\cite%
{afchar2018mesonet,bayar2016deep,cozzolino2017recasting,fridrich2012rich,rahmouni2017distinguishing}%
. We note, however, that the third scenario differs from the first two, as
it does not involve a change in identity. Our goal is to capture facial
manipulation cues caused by face swapping, where the apparent identity is
changed. Application-wise, swapping is of particular interest,
as many of the existing face manipulation methods are designed for
such identity modifying use cases. To this end we make two assumptions: (A1) Facial manipulation methods only manipulate the internal part of the face. (A2) The context of the face, which includes the head, neck, and hair regions outside the internal part of the face, provides a significant identity signal for the subject.

We verify assumption A2 in Sec.~\ref{sec:recnets}. Our findings are consistent with previous reports, showing that context alone indeed provides strong identity cues~\cite{kumar2009attribute,nirkin2018face}. 

To support assumption A1, Fig.~\ref{fig:manipulated_regions} visualizes the affected regions of six different state of the art facial manipulation methods. Fig.~\ref{fig:manipulated_regions}(a,b) present two reenactment methods by Thies et al.~\cite{thies2019deferred,thies2016face2face}. Both methods manipulate the regions corresponding to a 3D morphable model (3DMM)~\cite%
{blanz2002face,blanz2003face}, covering a facial region that contains part of the forehead at the top and most of the jaw on the bottom. Fig.~\ref{fig:manipulated_regions}(c,d) shows two deepfakes variants samples from the FaceForensics++~\cite{roessler2019faceforensics++} and DFD~\cite{dfd} datasets, both affecting a square region in the middle of the face. Fig.~\ref{fig:manipulated_regions}(e) is another 3DMM-based face swapping method, affecting similar regions as the reenactment methods, excluding the internal part of the mouth (sample obtained from previous work~\cite{roessler2019faceforensics++}). Fig.~\ref{fig:manipulated_regions}(f) is the output of FSGAN~\cite{nirkin2019fsgan} which uses face segmentation to manipulate entire face regions. 

We claim that it is no coincidence that all face manipulation methods we know of do not affect the entire head: While human faces have simple, easily modeled geometries, their context (neck, ears, hair, etc.) are highly irregular and therefore difficult to consistently reconstruct and manipulate, especially when considering the temporal constraints in video.

We present a novel signal for identifying fake images based on comparing the inner face region -- the one that is directly manipulated
-- with its outer context, which is left unaltered by all face manipulation methods we are aware of.
We do this by representing these two regions, faces and their context, with two separate
identity vectors. The two vectors are obtained by training two separate face
recognition networks: one trained for identifying a person based on
the face region and the other trained to identify the person based on face
context. We compare these two vectors, seeking
identity-to-identify discrepancies.

Importantly, we \emph{do not} assume prior knowledge of the identity of the
person appearing in the image (source or target subject identities).
Instead, given an image, we compare the representations for the one or two
(unknown) identities, obtained from the face and its context using our two,
specially trained networks.

The cue we derive using these two networks differs from those obtained by
methods that search for artifacts caused by particular face manipulation
techniques. Compared to other methods, our cue has three distinct advantages:
First, our cue is based on the inherent design of face swap
schemes and so is expected to hold even if future approaches produce
photo-realistic, artifact-free results. Second, this cue generalizes well to
different manipulation methods, whereas artifact detecting methods rely on
algorithm-specific flaws. Finally, since the proposed cue is largely unrelated
to artifact detection methods, it is complementary, and can thus be readily
combined with such approaches to improve accuracy.

To summarize, we make the following contributions: (1) We propose a novel approach to identifying the results of face swapping methods. (2) Our method is based on a novel fake detection cue that compares two image-derived identity {embeddings}. (3) The proposed approach is shown to outperform existing state-of-the-art schemes when applied to FaceForensics++~\cite{roessler2019faceforensics++}, Celeb-DF-v2~\cite{li2019celeb}, and DFDC~\cite{dolhansky2019deepfake}. (4) We show further results on two additional face swapping benchmarks, created using the FaceForensics++ data and additional swapping techniques, not included in FaceForensics++.

\section{Related work}
\label{sec:related_work} 

\noindent{\textbf{Face swapping techniques}.} Semi- and fully-automatic
face swapping methods were introduced nearly two decades ago~\cite{bitouk2008face,blanz2004exchanging}. These early methods were proposed as a means for preserving privacy~\cite{blanz2004exchanging,lin2012face,mosaddegh2014photorealistic}, recreation~\cite{kemelmacher2016transfiguring}, and entertainment (e.g.,~\cite{alexander2009creating,wolf2010eye}); a far cry from some of their less appealing applications today in misinformation and fake news. Nearly all pre-deep learning approaches relied to some extent on 3D face
representations, notably 3DMM~\cite%
{blanz2002face,blanz2003face,chang2019deep}. Some of the more recent examples of such
methods are the Face2Face approach for expression transfer~\cite%
{thies2016face2face}, face reenactement \cite{suwajanakorn2017synthesizing},
expression manipulation~\cite{masi2016we,masi2019face,averbuch2017bringing}, and face
swapping~methods~\cite{nirkin2018face}. 

Public awareness of face manipulation methods began following the introduction of deep
learning--based swapping and reenactment, particularly through the use of generative
adversarial networks (GAN). A few notable examples of such techniques are
GANimation~\cite{pumarola2018ganimation}, GANnotation~\cite%
{sanchez2018triple}, and others~\cite%
{kim2018deep,natsume18fsnet,natsume2018rsgan}. Unlike earlier,
3D-based methods, GAN-based approaches are able to produce near
photo-realistic results, not only in still photos, but also in videos. The
quality of these results, along with the availability of public
software, led to the use of what is now collectively known as \emph{DeepFakes}, 
for undesirable applications, including porn and fake news. More recently, FSGAN~\cite{nirkin2019fsgan} showed convincing swapping results without requiring a dedicated training procedure for each source or target person, i.e., it is trained to replace any face with any other face.

\subsection{Detecting manipulated faces} Over the years, many proposed methods for detecting generic, copy-move and splicing manipulations in images and videos~\cite{jia2018coarse,wu2018busternet,wu2018image,wu2019mantra}. Faces, however, received far less attention, likely because until recently, it was far harder to produce photo-realistic face manipulations. 

The elevated threat posed by recent face manipulation methods is now being answered by increased efforts to develop automatic fake detection methods. Early methods for detecting manipulated visual media relied on handcrafted features~\cite{fridrich2012rich}. A more modern, deep learning--based implementation of this approach was recently described by Cozzolino et al.~\cite{cozzolino2017recasting}, followed by other deep learning--based methods,~\cite{afchar2018mesonet,bayar2016deep,korshunov2018speaker,li2018ictu,li2018exposing,quan2018distinguishing,rahmouni2017distinguishing,rossler2018faceforensics,sabir2019recurrent}, as well as approaches utilizing multiple cues \cite{matern2019exploiting,nguyen2018modular,nguyen2019use,wang2019detecting,wu2019mantra,yang2019exposing,zhou2017two}.

Sabir et al.~\cite{sabir2019recurrent} recently proposed a recurrent
neural network which uses temporal cues to detect Deepfake
manipulations in videos. Stehouwer et al.~\cite{stehouwer2019detection}
applied an attention mechanism to intermediate feature maps of different
backbone classifiers, to improve manipulated region detection accuracy. Songsri et al.~\cite{songsri2019complement} showed that
using additional facial landmarks improves both detection and
localization of Deepfakes. Finally, Nguyen et al.~\cite{nguyen2019use} suggested a fake detection architecture based on the
capsule networks. Their work achieves results equivalent to previous methods, while 
utilizing significantly fewer parameters.

\subsection{Benchmarking face manipulation} A number of recent efforts try to provide the research community with standard,
high quality, fake detection benchmarks. These efforts include FaceForensics~%
\cite{rossler2018faceforensics}, DeepFake-TIMIT~\cite%
{korshunov2019vulnerability}, Celeb-DF~\cite{li2019celeb}, VTD dataset~\cite%
{al2016development}, FaceForensics++ challenge~\cite%
{roessler2019faceforensics++}, and the DFD dataset~\cite{dfd}. Several industry research labs have also
recently contributed to these efforts, leading to the announcement of the
DeepFake Detection Challenge (DFDC)~\cite{dolhansky2019deepfake}.

These benchmarks represent multiple manipulation techniques -- not just face
swapping. By using a single (or few) synthesis methods, biases can be
inadvertently introduced into these challenges: artifacts that are unique to
a particular fake generation method, or to the use of particular training
data. These sets, therefore, include media generated with a variety of
synthesis methods. Our approach is designed to be invariant to such
incidental biases: Rather than seeking particular artifacts, we consider a
perceptual effect shared by swapping techniques in general and show that our method can detect fakes produced by
previously unseen face manipulation techniques.

\section{Recognition of faces and their context}
\label{sec:faces}
We describe the two complementary face recognition networks used to obtain
identity cues for the face and its context. We further explain how we use these two networks in our proposed fake detection method.
Deep neural networks are extensively used for face identification, and we focus on the contributions of two very specific facial regions, dictated by the desired application: the segmented face and its surrounding context.

\subsection{Detecting and segmenting faces}
We begin by applying the dual shot face detector (DSFD)~\cite{li2019dsfd}. We then increase detected bounding box sizes by 20\%, relative to their height, to expose more of the context around the face, as DSFD is trained to return tight facial bounding boxes. Face crops are then resized to 299$\times $299 pixels; the input
resolution of the Xception architecture~\cite{chollet2017xception} which we use for
our face/context cues (Sec.~\ref{sec:recnets}).

To determine which parts of the crop are processed by the face network and which by the context network, we segment the crop into foreground (face) and background (context) using a face segmentation network. 
The exact architecture and training details for the segmentation network are provided in Appendix~\ref{sec:segmentation}.
Given the cropped face $I$ and its corresponding face segmentation mask $S$, we
generate image $I_{f}$ and its complementary image $I_{c}$, representing the face and its context, respectively.

\begin{figure*}[t]
\centering
\begin{subfigure}{.5\textwidth}
  \centering
  \includegraphics[width=0.95\linewidth]{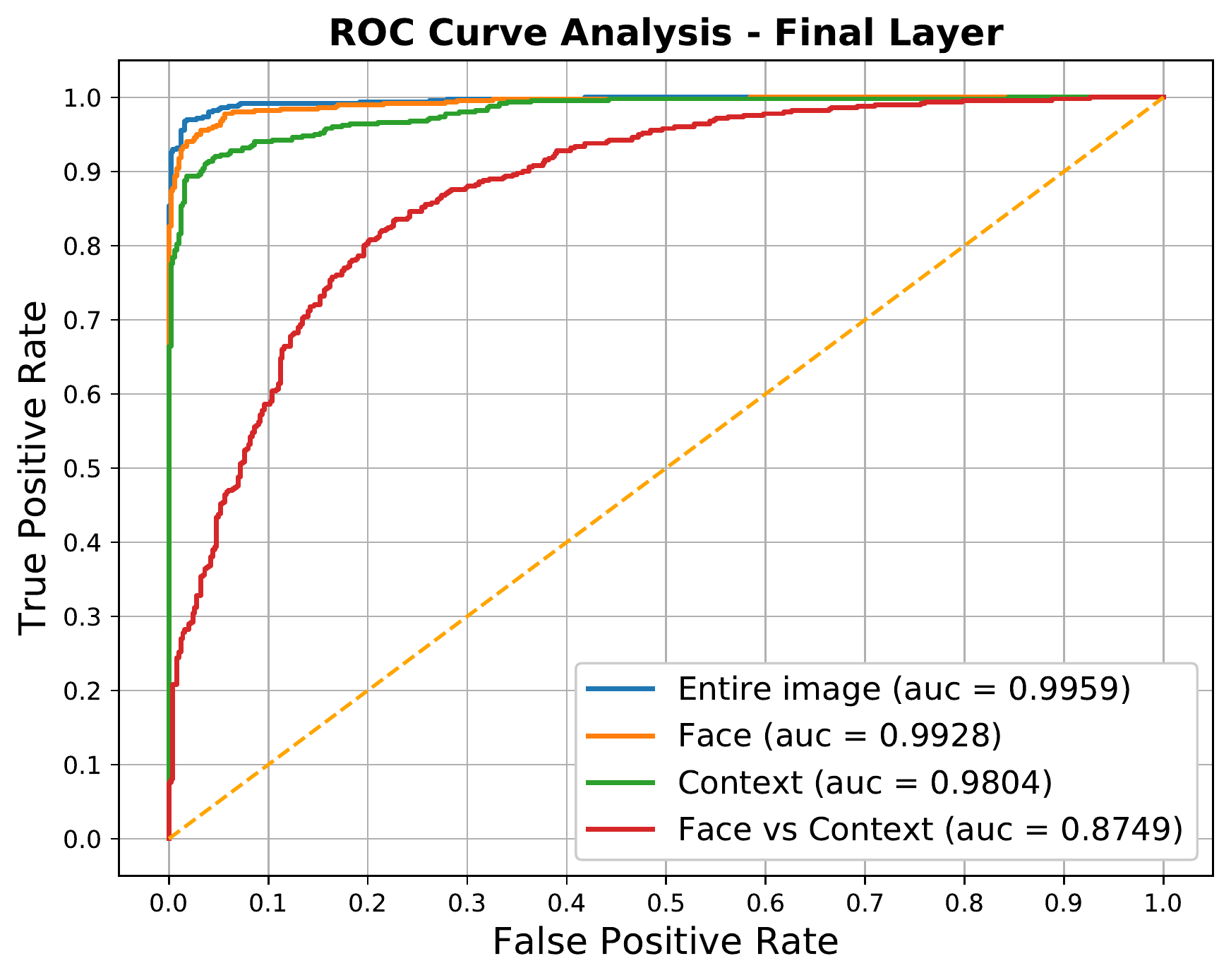}
  \caption{}
\end{subfigure}%
\begin{subfigure}{.5\textwidth}
  \centering
  \includegraphics[width=0.95\linewidth]{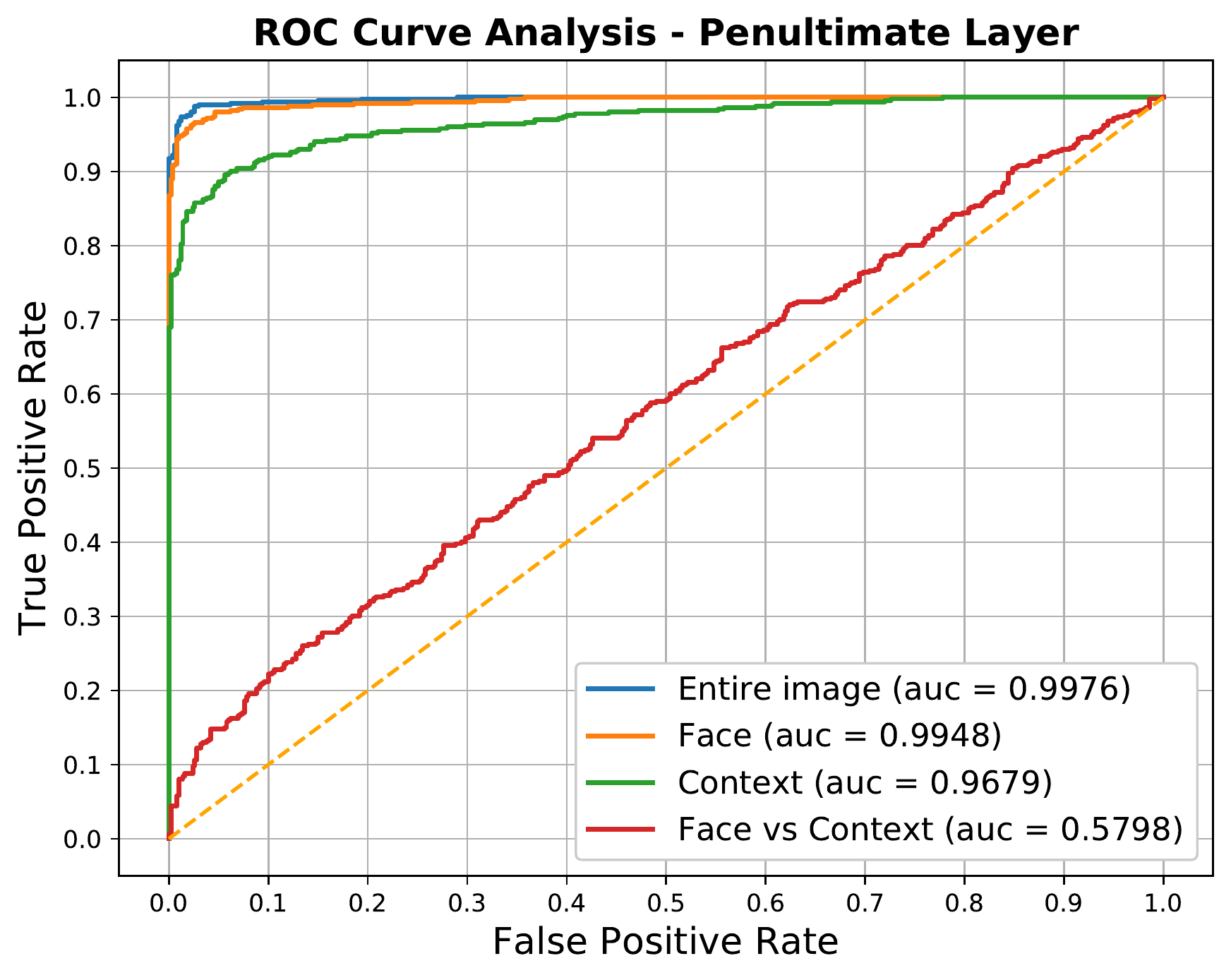}
  \caption{}
\end{subfigure}
\caption{{\bf LFW verification accuracy for identification networks trained on different face regions.} (a) Results obtained by representing faces with the final layers of the Xception architectures. (b) Faces represented using the activations of the penultimate layers of Xception. In the latter case, face vs. context do not match well for the same person, since the two networks were trained independently. Our approach, therefore, uses the final layers of the networks, representing subject pseudo-probabilities, when comparing the two (top).}
\label{fig:test}
\label{fig:roclfw}
\end{figure*}

\subsection{Recognition networks}\label{sec:recnets}
\label{sec:recog}

\noindent \textbf{Recognition network architecture.} Our networks are based on the Xception architecture~\cite{chollet2017xception} following its success in detecting other DeepFake cues~\cite{roessler2019faceforensics++}. We train the network using a vanilla cross entropy loss, although other loss functions could
presumably also be used. Xception is based on the Inception architecture~\cite{szegedy2017inception} but with Inception modules replaced with
depth-wise separable convolutions. As far as we know, it was
never used for face recognition.

In our implementation, the Xception network consists of a strided
convolution block, followed by twelve depth-wise separable
convolutions blocks with residual connections, except for the last one. The
network is terminated by two depth-wise separable convolutions, a
pooling operation and a fully connected layer.

We train two identification networks: $E_{f}$ which maps an image of size
299x299 containing pixels from the face region to a vector of
pseudo-probabilities associated with the dataset faces, and, similarly,
network $E_{c}$ maps the remaining pixels from the detection
bounding box (the context) to a vector of pseudo-probabilities of the same classes.

We train both $E_{f}$ and $E_{c}$ on images from the standard, publicly available VGGFace2 dataset~\cite{cao2018vggface2}. VGGFace2 contains 9,131 subjects from which we filtered images with a resolution lower than 128x128, resulting in 8,631 identities. The output of these two networks is, therefore, in $\mathbb{R}^{8,631}$.

\minisection{Validating recognition capabilities} To validate and compare the recognition accuracy of these networks, we test their performance on both the VGGFace2~\cite{cao2018vggface2} test set and the test set of the Labeled Faces in the Wild (LFW)~\cite{LFWTech} benchmark (no additional training or fine tuning was applied to the networks before being tested on LFW images).

Unsurprisingly, addressing the internal appearance of the face, network $%
E_{f}$ outperforms $E_{c}$ in term of accuracy, though both accuracies are high. These results
are evident from Table~\ref{tab:resultsonvggtestset} for VGGFace2 and Fig.~%
\ref{fig:roclfw} for LFW. We note that the accuracy demonstrated by $%
E_{c}$ -- its ability to recognize faces despite only seeing the context --
is unsurprising: similar results were reported by others,
showing that faces can be recognized, even when only their context is
visible~\cite{kumar2009attribute,nirkin2018face}.

Importantly, Fig.~\ref{fig:roclfw}(b) shows that the representations
typically used for face recognition -- the activations of the penultimate
layer of the face recognition network, do not match well for the same
person, since the two networks were trained independently. When combining the
responses from these two networks, we, therefore, use their final output: the
per-subject pseudo-probabilities (Sec.~\ref{sec:discrepancy}).

\begin{table}[t]
\centering
\begin{tabular}{lcc}
\toprule Method & Train set & Validation set \\[0.5ex]
\midrule Context & 99.90 & 87.06 \\
Face & 99.89 & 95.10 \\
Entire region & \textbf{99.98} & \textbf{96.98} \\
\bottomrule 
\end{tabular}
\caption{\textbf{Face recognition accuracy on VGGFace2.} Results reported
for three face identification Xception networks, each applied to a
different part of the face. As expected, the entire region, containing both
face and context, is the most accurate. Even context alone, however, provides a strong cue for identification, as previously observed by others~\cite{kumar2009attribute,nirkin2018face}. \vspace{-4mm}}
\label{tab:resultsonvggtestset}
\end{table}

\section{Fake detection using faces vs. context}

\begin{figure*}[t]
\centering
\includegraphics[width=\linewidth]{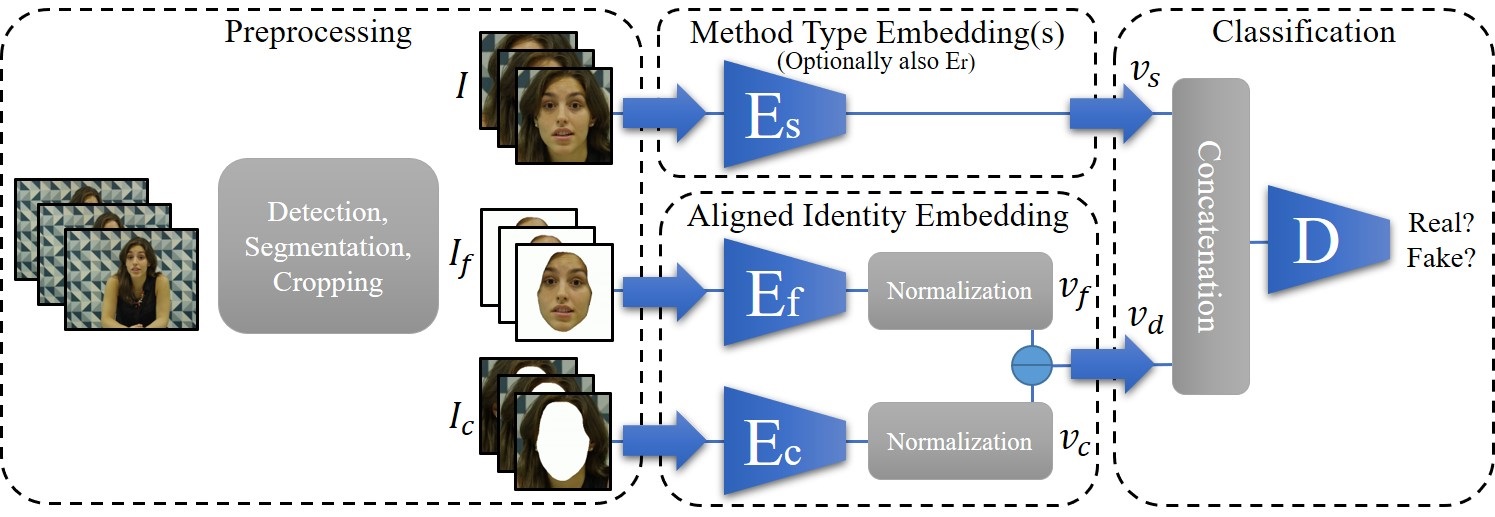}
\caption{\textbf{Method overview}. 
Following initial preprocessing,
we obtain regions for the face, $I_f$, and its context, $I_c$. The two are
processed by the face identification networks, $E_f$ and $E_c$,
respectively. A separate network, $E_s$, considers the input image, $I$, seeking apparent swapping artifacts to
decides if it is a face swapping result.
The pseudo-probability vectors of the two face identification networks are
subtracted and, jointly with the representations obtained from the method type network, $E_s$, are passed to the final
classifier, $D$. \vspace{-4mm}}

\label{fig:system}
\end{figure*}
We illustrate our proposed fake detection approach in Fig.~\ref%
{fig:system}. Our method combines multiple Xception networks: The recognition
networks, $E_{f}$ and $E_{c}$, described in Sec.~\ref{sec:faces}, a binary
Xception net, $E_{s}$, trained to distinguish between real and manipulated
images by face swapping methods, and another, {\em optional}, binary Xception net,
$E_{r}$ (not shown in Fig.~\ref{fig:system}), which we train to differentiate real images from those manipulated
by face reenactment methods. We next describe these components in detail.

\subsection{Face discrepancy component}
\label{sec:discrepancy} We train the face discrepancy network to predict whether a face and its context share the same identity. It uses the output of the two recognition networks, $E_{f}$ and $E_{c}$, described in Sec.~\ref{sec:faces}. We pre-train these two networks and do not change their weights after they are combined, in order to ensure that the identity cues remain the dominant ones.
In Sec.~\ref{sec:general} we show that training with the recognition network's weights unfrozen leads to a reduced accuracy when generalizing to unseen methods.

We process the face and context images, $I_{f}$ and $I_{c}$, with two separate identity classifiers, $E_{f}$ and $E_{c}$, respectively, to compute a discrepancy feature vector $v_{d}$
\begin{equation}v_{d}=E_{f}(I_{f})-E_{c}(I_{c})=v_{f}-v_{c}.
\label{eq:vd}
\end{equation}

\subsection{Manipulation specific networks}
\label{sec:specific} Previous approaches trained classifiers to
distinguish between real and fake faces, without considering the particular manipulation applied to the faces -- swapping or reenactment. These two manipulations types differ significantly: Swapping manipulates the identity of the face, whereas reenactment manipulates facial pose and expression. While the latter is not the focus of our work, it is required by the FaceForensics++ benchmark used in our tests (Sec.~\ref{sec:ffpp}). Our approach, therefore, includes also a component for
detecting face reenactment.

Specifically, we decouple swapping and reenactment by training a separate,
dedicated classifier for each: Network $E_{s}$ is trained to detect swapping
artifacts and network $E_{r}$ (not shown in Fig.~\ref{fig:system}) is trained to detect reenactment. We use
Xception networks, similar to those described in Sec.~\ref{sec:recnets} for
recognition, and train these networks to classify genuine vs. manipulated. Our training process first pre-trains both networks on examples of their particular manipulation vs. pristine images.
Our reenactment network, $E_{r}$, is used in cases where the task is to detect both face swapping and face reenactment methods. Otherwise, we use a three network solution, where $E_{r}$ is omitted.

\subsection{Combining all detection cues}
We chose the simplest method for combining the various signals: concatenating the three vectors $v_{d}$, $v_{s}$ and $v_{r}$, where $v_{d}\in \mathbb{R}^{8,631}$ is defined in Eq.~\eqref{eq:vd}, and $v_{s}=E_{s}^{p}(I)$ and $v_{r}=E_{r}^{p}(I)$, both in $\mathbb{R}^{2,048}$, denote the activations of the penultimate layers of the binary $E_{s}$ and $E_{r}$, respectively. 

The concatenated vector is passed to classifier $D$, which outputs a real vs. fake binary signal, trained using a logistic loss function. The classifier $D$ consists of an initial linear layer, followed by batch normalization, ReLU, and a final linear layer.

\subsection{Training}
We first pre-train the four classifiers, $E_{s}$, $E_r$, $E_{f}$, and $E_{c}$, each on its own task. We train network $E_{s}$ on the subset of videos in FaceForensics++~\cite{roessler2019faceforensics++} consisting of pristine videos and videos manipulated by the face swapping methods: FaceSwap and Deepfakes. Network $E_{r}$ is trained on the face reenactment methods: Face2Face and NeuralTextures. Note that we only use the compressed versions of these videos for training, with C23 (HQ) and C40 (LQ) compressions. We chose not to use the raw videos for training because there is little difference between them and the C23 compressed videos. The FaceForensics++ benchmark used to test our method does contain all three versions. 
The training process applied to $E_{f}$ and $E_{c}$ is detailed in Sec.~\ref{sec:faces}.

Once the four networks are trained, we freeze the weights of $E_{f}$ and $E_{c}$, and train the final classification network, $D$, using the three output vectors ($v_{s},v_{r},v_{d}$), while only fine-tuning $E_{r}$ and $E_{s}$. The final training is done on the same split of the FaceForensics++ videos. For more technical details, please see Appendix~\ref{sec:training_details}.

\subsection{Inference on full images}
During inference, we often process images containing multiple faces. In such cases, we only classify detected faces having a height larger than 64 pixels, and discard the rest as background faces. The only exceptions are images where the largest face does not comply with this criterion, in which case we process the largest detected face.

We further remove false detections by applying a threshold on the number of
face pixels in the face segmentation mask, $S$, for each detection. We start
with a threshold of 15\% of the face pixels, relative to the number of pixels in
the cropped region. If this step filters-out all our detections, we reduce
the threshold by half. If none of the images pass the 7.5\% threshold, we
simply consider the one face patch with the maximal number of detected
pixels.

Finally, we apply the compound network, including $E_{m},E_{f},E_{c}$, and $D
$, to the remaining face patches (one or more) and obtain one score per face
patch as the output of $D$. We take the minimal output of these scores --
the face patch predicted as most likely to be fake -- in cases where
only a single face is manipulated.

\section{Experimental results}\label{sec:experiments}

We evaluated our proposed scheme using three recent, challenging benchmarks: FaceForensics++~\cite{roessler2019faceforensics++}, DFDC~\cite{dolhansky2019deepfake}, and Celeb-DF-v2~\cite{li2019celeb}. In order to evaluate our method using additional face swapping techniques and test its generalization abilities, we further create our own test set, using two more swapping methods. For a runtime analysis of our method, please see Appendix~\ref{sec:runtime_performance}.

\subsection{Face swapping detection experiments}
\label{sec:face_swap_detection_experiments}
We use the following three datasets containing only face swapping examples:

\minisection{FF-DF} FF-DF~\cite{li2019celeb} is a subset of the FaceForensics++ benchmark~\cite{roessler2019faceforensics++}, which includes only faces swapped using the Deepfakes method~\cite{DeepFakes}. These tests therefore include 1,000 videos from the {\em pristine} subset and 1,000 videos from the {\em Deepfakes} subset (the full FaceForensics++ is described in Sec.~\ref{sec:ffpp}). 

\minisection{DFDC} The recently announced, industry-backed, preview of the DFDC benchmark~\cite{dolhansky2019deepfake} offers a total of 5,244 videos of 66 actors: 4,464 training videos and 780 test videos, 1,131 of them are real videos and 4,113 are fakes generated by two different, unknown, face swapping methods.

\minisection{Celeb-DF-v2} Another recent dataset containing 590 real videos and 5,639 DeepFake videos of 59 celebrities~\cite{li2019celeb}. This set is especially challenging as most state of the art methods tested on this set report near-chance accuracies.

\minisection{Training and evaluation} In these tests, we do not use our reenactment network, $E_r$. We train on FaceForensics++, as described in Sec.~\ref{sec:faces}. Results for all baseline  methods were previously reported~\cite{li2019celeb}. These methods were trained mainly on FaceForensics++, sometimes with additional self collected data. None of these methods was trained on DFDC or Celeb-DF-v2 and so these experiments also compare the generalization of the different methods.

All methods were compared using the area under the curve (AUC), at the frame level, on all frames in which faces were detected. This metric is very convenient for comparing methods that output per-frame classification as there is no need to set thresholds. 

\minisection{Face swap detection results}
We report our results in Table~\ref{tab:face_swap_detection_results}. Our method achieves the best AUC scores on all the benchmarks. On FaceForensics's DeepFakes subset~\cite{roessler2019faceforensics++} our method achieves similar results as the current state of the art, this is due to the accuracy being saturated. On the more challenging Celeb-DF-v2 benchmark, small improvements on the AUC scores are significant. Note also that the results reported for our method on Celeb-DF-v2 testify to its improved generalization abilities compared to the baseline methods.  

\begin{table}[t!]
\centering{ 
\begin{tabular}{@{}l@{~}ccc@{}}
\toprule Methods & FF-DF & Celeb-DF-v2 \\[0.5ex]
\midrule
Two-stream~\cite{zhou2017two} & 70.1 & 53.8 \\
\midrule
Meso4~\cite{afchar2018mesonet} & 84.7 & 54.8 \\
MesoInception4~\cite{afchar2018mesonet} & 83.0 & 53.6 \\
\midrule
HeadPose~\cite{yang2019exposing} & 47.3 & 54.6 \\
\midrule
FWA~\cite{li2018exposing} & 80.1 & 56.9 \\
DSP-FWA~\cite{li2018exposing} & 93.0 & 64.0 \\
\midrule
VA-MLP~\cite{matern2019exploiting} & 66.4 & 55.0 \\
VA-LogReg~\cite{matern2019exploiting} & 78.0 & 55.1 \\
\midrule
XceptionNet-raw~\cite{roessler2019faceforensics++} & \textbf{99.7} & 48.2 \\
XceptionNet-c23~\cite{roessler2019faceforensics++} & \textbf{99.7} & 65.3 \\
XceptionNet-c40~\cite{roessler2019faceforensics++} & 95.5 & 65.5 \\
\midrule
Multi-task~\cite{nguyen2019multi} & 76.3 & 54.3 \\
\midrule
Capsule~\cite{nguyen2019use} & 96.6 & 57.5 \\
\midrule 
Ours & \textbf{99.7} & \textbf{66.0} \\
\bottomrule
\end{tabular}
}
\caption{{\bf Face swap detection results.} Comparison of our approach and leading state of the art methods on two benchmarks using frame-level AUC~(\%).\vspace{-3mm}}
\label{tab:face_swap_detection_results}
\end{table}

\subsection{Experiments on FaceForensics++}
\label{sec:ffpp}
The full FaceForensics++ dataset~\cite{roessler2019faceforensics++} contains 1,000 videos obtained from the web, from which 1,000 video pairs were randomly selected and used to generate additional 1,000 manipulated videos representing four face manipulation schemes. Two of these methods perform face swapping: a 3D-based face swapping method~\cite{FaceSwap} using a traditional graphics pipeline and blending, and a GAN-based method~\cite{DeepFakes}, trained using the images of pairs of subjects to compute a mapping between them. Two additional methods perform face reenactment: Face2Face~\cite{thies2016face2face}, a 3DMM-based method that manipulates facial expressions by changing the expression-coefficients estimated for the face, and NeuralTextures~\cite{thies2019deferred} which learns a face neural texture from a video and uses it to realistically render a 3D reconstructed face model.

\minisection{Results on FaceForensics++ image benchmark} Table~\ref{tab:ff_image_results} shows that our total accuracy outperforms all previous methods by a large margin. Importantly, the accuracy in each of the different categories, on its own, is not a direct indication of detection performance, since there is a threshold-dependent trade-off between the accuracy on real and fake images. These results hint at the relative detection difficulty of each class and are provided for completeness. 

\begin{table}[t!]
\centering{
\begin{tabular}{@{}l@{~}cccccc@{}}
\toprule Methods & DF & F2F & FS & NT & Pristine & Total \\[0.5ex]
\midrule Steg. Features~\cite{fridrich2012rich} & 73.6 & 73.7 & 68.9 & 63.3 & 34.0 & 51.8 \\
Cozzolino et al.~\cite{cozzolino2017recasting} & 85.4 & 67.8 & 73.7 & 78.0 & 34.4 & 55.2 \\
Rahmouni et al.~\cite{rahmouni2017distinguishing} & 85.4 & 64.2 & 56.3 & 60.0 & 50.0 & 58.1 \\
Bayar and Stamm~\cite{bayar2016deep} & 84.5 & 73.7 & 82.5 & 70.6 & 46.2 & 61.6 \\
MesoNet~\cite{afchar2018mesonet} & 87.2 & 56.2 & 61.1 & 40.6 & 72.6 & 66.0 \\
Xception~\cite{roessler2019faceforensics++} & 96.3 & 86.8 & 90.3 & 80.7 & 52.4 & 71.0 \\
\midrule
Ours & 94.5 & 80.3 & 84.5 & 74.0 & 67.6 & \textbf{75.0} \\
\bottomrule
\end{tabular}
}
\caption{{\bf FaceForensics++ image benchmark results.} Columns are: {\em DeepFakes} (DF), {\em Face2Face} (F2F), {\em FaceSwap} (FS), {\em NeuralTextures} (NT), and {\em Pristine} categories. It is hard to compare specific columns, since there is a threshold-based trade-off between real and fake. These columns are therefore provided only for completeness. Our method leads in the Total score, which is the meaningful metric for this benchmark.\vspace{-3mm}}
\label{tab:ff_image_results}
\end{table}

\subsection{Ablation study and generalization experiment}
\label{sec:general}

\begin{figure*}[t]
\centering
\includegraphics[width=1.0\linewidth]{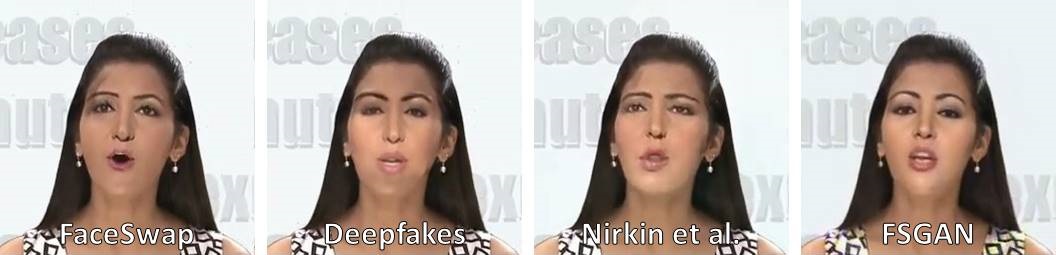}
\caption{{\bf Extending FaceForensics++ with unseen methods.} Examples shown for the same source / target face pair, using the 3D-based methods, FaceSwap~\cite{FaceSwap} and Nirkin et al.~\cite{nirkin2018face}, and the GAN-based methods, Deepfakes~\cite{DeepFakes} and FSGAN~\cite{nirkin2019fsgan}. Despite using the same image pairs in all four examples, the results are different, each exhibiting its own artifacts.}\label{fig:augmenting_qualitative}
\end{figure*}

\begin{figure*}[t]
\centering
\begin{subfigure}{.5\textwidth}
  \centering
  \includegraphics[width=.95\linewidth]{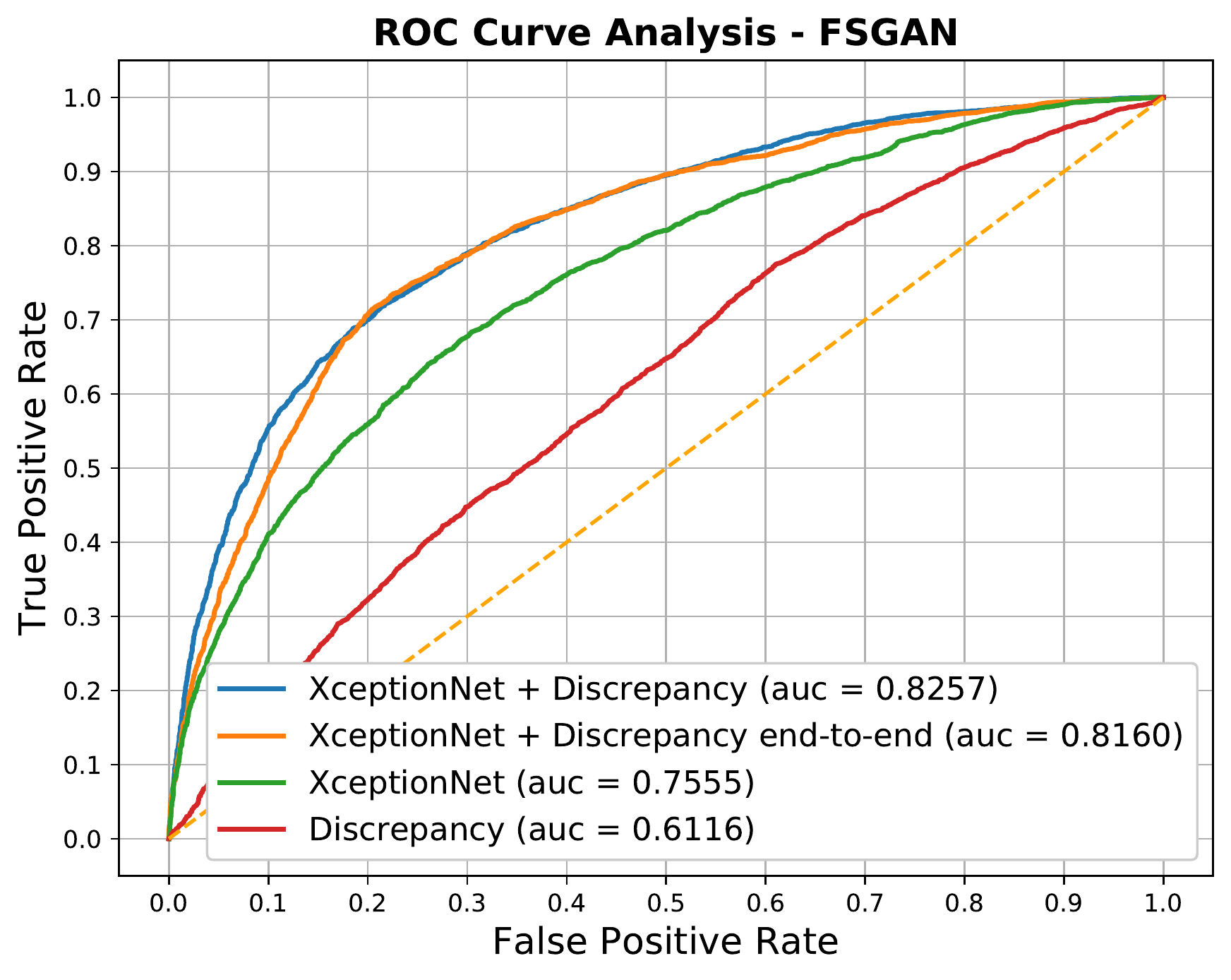}
  \caption{}
\end{subfigure}%
\begin{subfigure}{.5\textwidth}
  \centering
  \includegraphics[width=.95\linewidth]{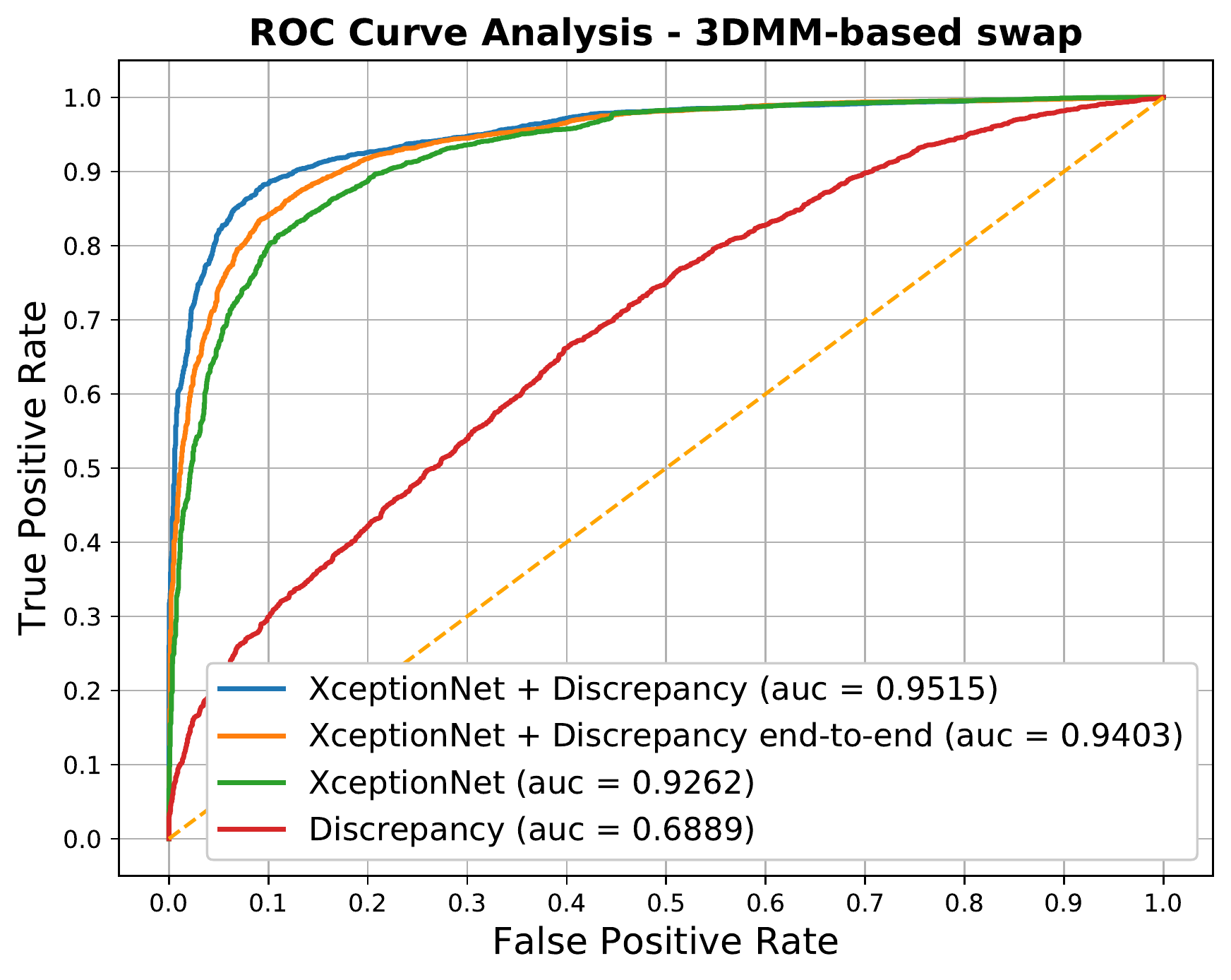}
  \caption{}
\end{subfigure}
\caption{{\bf Results on our two variations of FaceForensics++ videos.} (a) Generalization results with FSGAN generated swaps~\protect\cite{nirkin2019fsgan}. (b) Generalization results with swaps generated by Nirkin et al.~\cite{nirkin2018face}. See Sec.~\ref{sec:general} for more details. \vspace{-4mm}}
\label{fig:nirkin_roc}
\end{figure*}

\begin{figure*}[t]
\centering
\includegraphics[width=0.95\linewidth]{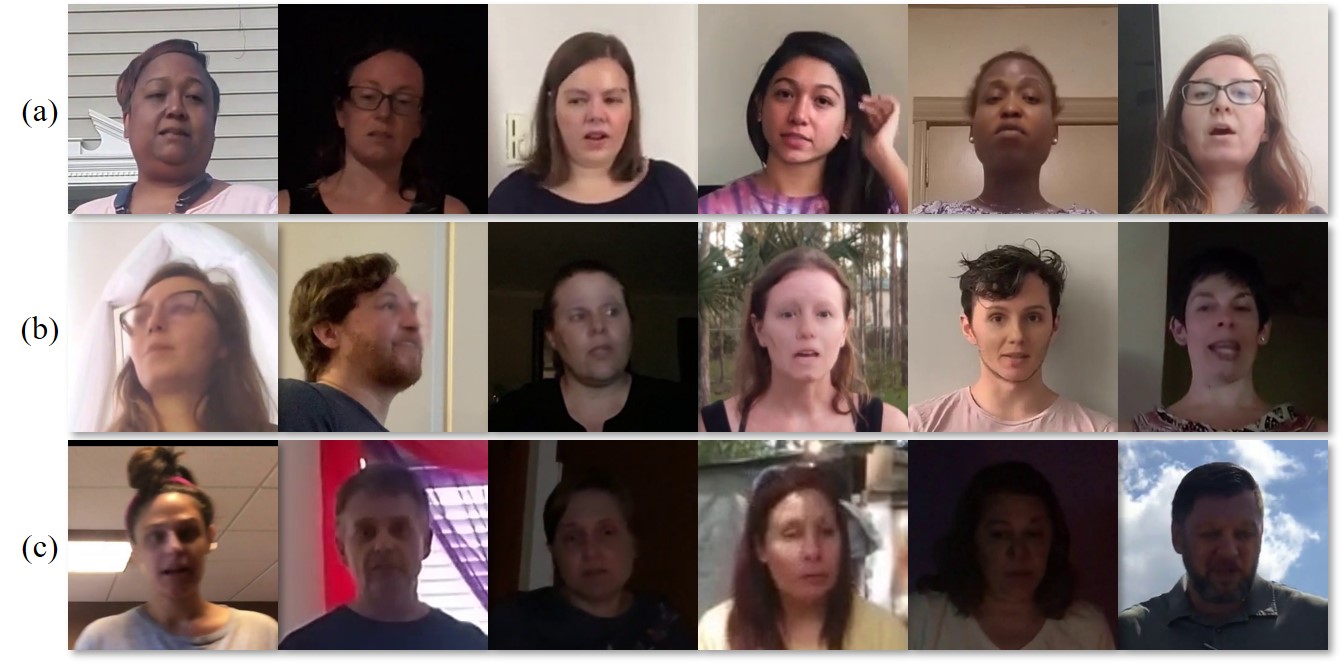}
\caption{{\bf Qualitative detection results.} Examples taken from the DFDC collection. (a) Fakes detected by our method, but undetected by a leading baseline, XceptionNet fake detector~\cite{chollet2017xception}. (b) Fakes detected by XceptionNet but missed by our approach. (c) Fakes missed by both methods. See Sec.~\ref{sec:qualitative} for more details. \vspace{-4mm}}
\label{fig:qualitative}
\end{figure*}

\begin{table}[t!]
\centering{\ 
\resizebox{0.98\linewidth}{!}{
{\begin{tabular}{lcccccc}
\toprule & \multicolumn{3}{c}{3D-based swap} & \multicolumn{3}{c}{FSGAN}
\\
\cmidrule(lr){2-4}
\cmidrule(lr){5-7}
Methods & Fake & Real & Total & Fake & Real & Total \\[0.5ex]
\midrule Face identity difference & 47.33 & 77.66 & 62.50 & 34.66 & 80.50 & 57.58 \\
Binary XceptionNet~\cite{cozzolino2017recasting} & 55.38 & \textbf{97.72} & 76.55 & 24.80 & 94.68 & 59.74 \\
Ours (end-to-end) & 54.74 & 97.70 & 76.22 & 31.66 & \textbf{95.38} & 63.52 \\
Ours & \textbf{68.20} & 95.10 & \textbf{81.65} & \textbf{47.14} & 90.56 & \textbf{68.85}
\\
\bottomrule

Face identity difference & 60.20 & 66.12 & 63.16 & 38.96 & \textbf{77.50} & 58.23 \\
Binary XceptionNet~\cite{cozzolino2017recasting} & 89.03 & 81.36 & 85.20 & 73.92 & 64.04 & 68.98 \\
Ours (end-to-end) & \textbf{90.77} & 83.54 & 87.16 & \textbf{79.58} & 71.40 & \textbf{75.49} \\
Ours & 90.52 & \textbf{88.20} & \textbf{89.36} & 78.72 & 71.66 & 75.19 \\
\bottomrule
\end{tabular}
}
}
}
\caption{{\bf Generalization results.} Top: Results with a fixed threshold at zero. Bottom: Upper bound results, obtained with a fixed threshold maximizing total accuracy on the test set. See Sec.~\ref{sec:general} for more details.\vspace{-7mm}}
\label{tab:internal}
\end{table}

Face manipulation methods sometimes leave behind artifacts, possibly imperceptible, that can be leveraged for detection. Different manipulation methods, however, can produce different artifacts, as shown in Fig.~\ref{fig:augmenting_qualitative}. There is, therefore, no guarantee that a fake detection method would perform well when presented with fakes generated by unseen schemes which do not leave such known, recognizable artifacts. We next verify the accuracy of our proposed scheme in detecting fakes produced by methods that were not part of its training set.   

We conduct these tests by extending the FaceForensics++ set, applying two additional face swapping methods to its videos: (1) Our implementation of FSGAN~\cite{nirkin2019fsgan} and (2) the publicly available implementation of Nirkin et al.~\cite{nirkin2018face}, a 3D-based face swapping method that uses single image 3D face reconstruction and segmentation. Examples of the four face swapping methods, using the same source and target, can be seen in Fig.~\ref{fig:augmenting_qualitative}. Each method generates face swaps with distinct artifacts, with the exception of FSGAN, which produces images with fewer apparent artifacts.

The extended version of the benchmark follows the pair selections prescribed by the original FaceForensics++ dataset. 
Because Nirkin et al.~\cite{nirkin2018face} was designed for image-to-image face swapping, for each frame in the target video we select its closest frame in the source video, in terms of estimated head pose.

In all our generalization experiments, we train the variants of our method and its XceptionNet baseline on the pristine and face swapping manipulations, using the official training and validation subsets of FaceForensics++. In these experiments, we do not use the reenactment detection network $E_r$.

\subsubsection{Generalization and ablation results} 
We studied the effect of our face vs. context discrepancy approach by comparing it to a naive classifier. We further test an end-to-end version of our method, where all the classifiers are unfrozen in the training process. We report our generalization results in Table~\ref{tab:internal} (ROC curves provided in Fig.~\ref{fig:nirkin_roc}). For results appearing at the top of Table~\ref{tab:internal}, we fix the thresholds for XceptionNet and our method at zero. In the bottom of Table~\ref{tab:internal} we optimize both thresholds on the test set. The threshold of the face identity difference in the first experiment is optimized using the VGGFace2 test set.

Our results show that our method significantly outperforms the baseline on both unseen methods. The performance gap is greater on FSGAN generated faces, where artifacts are more rare. Artifacts produced by the 3DMM-based method are more similar to the ones we encounter in other methods, and so the gap is smaller. As evident from the ROC curves in Fig.~\ref{fig:nirkin_roc}, the end-to-end version of our method is less able to generalize. This result is due to the end-to-end training process sullying the face and context classifiers roles for extracting aligned identity representations. 

Finally, note that the face discrepancy signal by itself is not competitive with networks trained to detect fakes. However, it is indicative of fake videos and its contribution to the overall method is seen by comparing our method with the baseline XceptionNet.

\subsection{Qualitative results}\label{sec:qualitative} Fig.~\ref{fig:qualitative} presents qualitative examples of detected and missed fake faces from the DFDC collection. Fig.~\ref{fig:qualitative}(a) shows example fakes detected by our method but undetected by the state of the art XceptionNet fake detector~\cite{chollet2017xception}. Fig.~\ref{fig:qualitative}(b) offers example fakes which were detected by XceptionNet, but were missed by our method. Finally, Fig.~\ref{fig:qualitative}(c) shows fakes missed by both approaches. 

Clearly, our method excels in cases where swapping artifacts are hard to detect (Fig.~\ref{fig:qualitative}(a)). Examining Fig.~\ref{fig:qualitative}(b) shows that fake images detected by XceptionNet often exhibit visible artifacts, which that method was optimized to detect. Our method includes a face swapping component, $E_s$ (Sec.~\ref{sec:specific}), trained to detect similar method-specific artifacts, but does not provide the same detection accuracy as the baseline when such artifacts are present. Our overall approach still outperforms the baseline by a wide margin, as reported in Sec.~\ref{sec:face_swap_detection_experiments} and~\ref{sec:ffpp}. Finally, the fakes missed by both methods are typically challenging images with low contrast or blurry features as in Fig.~\ref{fig:qualitative}(c).

\section{Conclusion}

While the ability to manipulate faces in images and video has increased
dramatically in the last few years, all recent methods follow similar
patterns. In this work, we propose a novel detection cue which utilizes the
commonalities of all recent face identity manipulation methods. It is
complementary to conventional real/fake classifiers and can be used
alongside them. Overcoming this approach would require a much broader
integration of the new identity into the image, making our contribution hard
to circumvent without additional technological breakthroughs. This is in
contrast to artifact detection methods, which are susceptible to the
constant progress in the visual quality of generated images. It is our hope
that by further analyzing the design principles of face swapping techniques,
additional methods of identifying fake images and videos would be
discovered, leading to effective mitigation of the societal risks of such
media.

\appendices

\section{Segmentation network details}
\label{sec:segmentation}
For the face segmentation network, we are using the U-Net~\cite{ronneberger2015u} architecture where the deconvolution layers used for the upsampling operations replaced with bilinear interpolation followed by a convolution. For this network, we crop and resize the images to a resolution of 256x256.

The network is trained on a face segmentation dataset, similar to the one used by Nirkin el al.~\cite{nirkin2018face}, and produced by us using their publicly available code\footnote{Available: \url{https://github.com/YuvalNirkin/face_video_segment}}. Training used a batch size of 48 and 40,000 iterations per epoch. To increase the robustness of the segmentation network, the following image augmentations are applied: random image rotations between -30 to 30 degrees, random color jittering (brightness, contrast, saturation, and hue), horizontal flip with probability 0.5, and gaussian blur with a kernel size of 5 and sigma of 1.1, which is applied with probability 0.5.

\section{Training details}
\label{sec:training_details}
Training used four NVIDIA Tesla P100 GPUs and an Intel Xeon CPU with 64 cores. We applied Adam optimization~\cite{kingma2014adam} ($\beta_{1}=0.5,\beta_{2}=0.999$). The learning rate was reduced by half every ten epochs, for a total of 90 epochs. The initial learning rate and the batch size differ between the training stages (see below).

\minisection{Identity networks} The identity networks, $E_{f}$ and $E_{c}$, are trained on images from the VGGFace2 dataset~\cite{cao2018vggface2}, cropped by the provided bounding boxes after they have been squared using the length of the longer of the two axes and their size extended by 20\%. The cropped images are then resized to a resolution of 299x299, which is the required input size for the Xception architecture~\cite{chollet2017xception}. As a form of augmentation, the images are horizontally flipped with a probability of 0.5. For the face network, $E_{f}$, we set the context pixels to a constant color, and for the context network, $E_{c}$, we set pixels corresponding to the face regions to a constant color. Both networks are trained with a batch size of 192 and an initial learning rate of 0.0002.

\minisection{Pretraining of the manipulation specific networks} Networks $E_{s}$ and $E_{r}$, are trained on the videos of FaceForensics++~\cite{roessler2019faceforensics++}, each on its own specific subset as described in Sec. 4.4 in the main paper. The networks are trained for 40,000 iteration per epoch, with a batch size of 96, and an initial learning rate of 0.0002. For each iteration a random frame from a random video is selected, for which a face was detected. The face crops are transformed in the same way as for the identity networks.

\minisection{Training of the full pipeline} In this final training, the networks $E_{s}$, $E_{r}$, and $D$ are trained, while the weights of the identity networks, $E_{f}$ and $E_{c}$, are frozen (except for the end-to-end ablation experiment, see Sec. 5.3 in the main paper). The discriminator $D$ is trained from scratch, its weights are randomly initialized using a normal distribution. A batch size of 64 is used and there are 20,000 iterations per epoch. The  initial learning rate is set to 0.0001.

\section{Runtime performance}
\label{sec:runtime_performance}

The system runtime performance was tested on a single NVIDIA Tesla V100 GPU with an Intel Xeon CPU with 8 cores. The entire pipeline was tested, including the segmentation and preprocessing but without the face detection step.
For face detection we are using the dual shot face detector (DSFD)~\cite{li2019dsfd} which was not optimized for run-time performance and was originally used for images and not videos. This detection step can potentially be replaced with a face tracking algorithm (many of which can run in real-time even on CPU~\cite{baltrusaitis2018openface}). Our full pipeline inference rate without the reenactment classifier, $E_r$, is fast, at 90.6fps. Runtime is 81.5fps with $E_r$.

\pagebreak

\ifCLASSOPTIONcaptionsoff
\newpage \fi

\bibliographystyle{IEEEtran}
\bibliography{main}

\end{document}